# Supporting Regularized Logistic Regression Privately and Efficiently

Wenfa Li[1], Hongzhe Liu[1], Peng Yang[1], Wei Xie[2,*]

[1] Beijing Key Laboratory of Information Service Engineering, Beijing Union University, Beijing, 100101, China

[2] Department of Electrical Engineering & Computer Science, Vanderbilt University, Nashville, TN 37232, USA

[*] To whom correspondence should be addressed: wei.xie@vanderbilt.edu

## Abstract

As one of the most popular statistical and machine learning models, logistic regression with regularization has found wide adoption in biomedicine, social sciences, information technology, and so on. These domains often involve data of human subjects that are contingent upon strict privacy regulations. Increasing concerns over data privacy make it more and more difficult to coordinate and conduct large-scale collaborative studies, which typically rely on cross-institution data sharing and joint analysis. Our work here focuses on safeguarding regularized logistic regression, a widely-used machine learning model in various disciplines while at the same time has not been investigated from a data security and privacy perspective. We consider a common use scenario of multi-institution collaborative studies, such as in the form of research consortia or networks as widely seen in genetics, epidemiology, social sciences, etc. To make our privacy-enhancing solution practical, we demonstrate a non-conventional and computationally efficient method leveraging distributing computing and strong cryptography to provide comprehensive protection over individual-level and summary data. Extensive empirical evaluation on several studies validated the privacy guarantees, efficiency and scalability of our proposal. We also discuss the practical implications of our solution for large-scale studies and applications from various disciplines, including genetic and biomedical studies, smart grid, network analysis, etc.

# Introduction

The ever-increasing amount of data have posed great demand for effective analytical methods to sift through them. Logistic regression and its regularized variants [1, 2] are among the most widely-used classification models in machine learning and statistics. It has seen a wide range of applications across various human endeavors, including genetics/genomics (e.g., genome-wide association studies, or GWAS [3], gene-gene interaction detection [4]), epidemiology (e.g., [5, 6]), social sciences [7, 8], information technology (e.g., computational advertising on the internet [9] and personalized recommender systems [10]), etc.

Many of the aforementioned disciplines and applications rely on huge numbers of data records (commonly referred to as large sample sizes in many fields) to make reliable discoveries or predictions. The scale of data desired is often beyond the capability of any single institution, and thus depends heavily on collaboration across different institutions through data collection, data sharing and collaborative analysis.

However, data sharing and collaborative studies across different institutions bring about serious privacy concerns, as most such studies involve raw data of human subjects that are considered private and sensitive. In biomedicine, for instance, individual patient records are highly sensitive and protected under stringent regulations such as the Health Insurance Portability and Accountability Act (HIPAA) [11]; Genetic information of humans are also deemed highly sensitive [12, 13] and partially covered by the Genetic Information Nondiscrimination Act (GINA) [14]; in the education domain, students' data privacy is strictly regulated under the Family Education Rights and Privacy Act (FERPA) [15]. In other domains, ignorance of data privacy and misuse of personal information has even outraged users [16] and raised awareness of regulators [17], as in the case of internet targeted advertising. Meanwhile, various high-profile data breaches [18, 19] have exacerbated the situation, damaging the credibility of centralized data hosts and analytical centers in upholding user privacy.

A classical approach to alleviating privacy concerns is by concealing individual raw data via artificial perturbation (e.g., *k*-anonymity [20] or differential privacy [21]), Cryptography-based methods (e.g., encrypting genetic data [22]), or distributed computing (e.g., private records reside at local institutions only [6, 13, 23]). Increasingly, such protections often prove to be insufficient, due to various privacy attacks [12, 24–27] leveraging numerous types of side channels (mostly aggregate information or summary statistics), such as allele frequencies from published GWAS studies and public reference genotypes of humans, correlation quantification between genetic variants in the form of linkage disequilibrium (LD), regression coefficients or effect size estimates, p-values, variance-covariance, etc.

Our work here studies the data privacy issues in regularized logistic regression [2]. Regularized logistic regression is widely used in various domains, and is often the preferred model of choice over standard logistic regression in practice [2, 4, 28, 29]. Despite its popularity, it has received little investigation from a data privacy and security perspective. Our work intends to bridge the gap.

In this work, we show how to perform regularized logistic regression while preserving data privacy. To do so, we adapt an efficient optimization method based on distributed computing [6]. The method partitions and distributes sensitive computations such that no (private) raw individual data need to be shared beyond their owner institutions. This leads to better privacy protection on raw data and orders-of-magnitude efficiency gains over a straightforward centralized implementation. In addition, we propose highly secure and flexible protocols to protect intermediate data and computations from regularized regression model fitting. These altogether lead to an efficient framework for safeguarding regularized logistic regression which provides comprehensive privacy protection over raw as well as intermediate data. We focus on use scenarios where multiple disparate institutions hope to collaboratively perform joint studies (ideally on their consolidated data collection), however they do not want to disclose their respective data to others due to privacy and/or confidentiality concerns. Such scenarios are ubiquitous in large collaborations in healthcare, genetics, epidemiology, finance, network analysis and so on (as we will elaborate later).

In summary, we consider our contributions to be three-fold:

- Firstly, we demonstrate that regularized logistic regression can be supported efficiently without violating privacy. As mentioned earlier, regularized logistic regression is widely used in practice and enjoys continued investigation from a methodological and computational perspective, yet very few efforts have been devoted to address its related privacy issues. Our work is the first to address such an important issue.

- Secondly, we present a secure and efficient method tailored for regularized logistic regression. We adapt an emerging method of distributed Newton-Raphson [6] for our problem of focus, enhance and extend its privacy protection leveraging strong cryptographic techniques [30]. Our resulting framework not only safeguards regularized logistic regression in particular, but is also relevant to the broader community of privacy-preserving regression analysis where intermediate data do not often receive sufficient protection.

- Lastly, we validate our privacy-enhanced regularized logistic regression extensively with both synthetic and real-world studies. We also demonstrate its scalability to large-scale collaborative studies, and illustrate its practical relevance to various applications from different disciplines.

## Materials and Methods

We first briefly introduce the necessary background regarding regularized logistic regression and the model estimation process. Later, we will show how such a process can be adapted to preserve data privacy without introducing significant computational overhead.

**Preliminaries**

**(Regularized) Logistic Regression.** Logistic regression [1] is a probabilistic classification model for predicting binary or categorical outcomes through a logistic function. It is widely used in many domains such as biomedicine [4, 5, 31], social sciences [7, 8], information technology [9, 10], and so on. Briefly, logistic regression is of the form:

$$p(y = 1|\mathbf{x}; \boldsymbol{\beta}) = \frac{1}{1 + e^{-\boldsymbol{\beta}^T \mathbf{x}}}, \tag{1}$$

where $p(.)$ denotes the probability of the response $y$ equal to 1 (i.e., "case" or "success" depending on the scenario), $\mathbf{x}$ is the $d$-dimensional covariates (or features) for a specific data record, and $\boldsymbol{\beta}$ is the regression coefficients we want to estimate.

Regularized logistic regression [2, 32] shares the same model as illustrated above. However, it differs in the model estimation process (with additional regularization terms added to the optimization objective), which leads to some desirable properties such as better model generalizability, support for feature selection, etc. As a result, regularized logistic regression is often the preferred choice for many real applications in practice [4, 33, 34].

In this work, we focus on regularized logistic regression with the $L_2$ norm [2], i.e., with the regularization term equal to $\frac{\lambda}{2}||\boldsymbol{\beta}||_2^2$, where $\lambda$ is the regularization parameter and $\boldsymbol{\beta}$ is the regression coefficients (note that incorporating other regularizations such as the $L_1$ norm [32] is also possible).

**Newton-Raphson Method.** A common way to estimate the (regularized) logistic regression model (i.e., to obtain $\boldsymbol{\beta}$ coefficients in Equation 1) is through the Newton-Raphson method (or the iteratively reweighted least squares method) [35, 36]. The repeated Newton-Raphson method adopts an iterative refinement process that eventually converges to the "true" values of the $\boldsymbol{\beta}$ coefficients.

To illustrate the process, we use $\boldsymbol{\beta}^{old}$ and $\boldsymbol{\beta}^{new}$ to denote the $\boldsymbol{\beta}$ coefficient estimates for the current and next iterations, respectively. Each step of the Newton-Raphson method can be expressed as:

$$\boldsymbol{\beta}^{new} = \boldsymbol{\beta}^{old} - \mathbf{H}^{-1}(\boldsymbol{\beta}^{old}) \, \mathbf{g}(\boldsymbol{\beta}^{old}), \tag{2}$$

where $\mathbf{H}(\boldsymbol{\beta}^{old})$ and $\mathbf{g}(\boldsymbol{\beta}^{old})$ denote the Hessian matrix and gradient of the objective function evaluated at the current estimate of the $\boldsymbol{\beta}$ coefficients. Details of computing $\mathbf{H}(.)$ and $\mathbf{g}(.)$ will be introduced later.

**Our Proposal**

Here, we introduce our privacy-preserving approach for supporting $L_2$-regularized logistic regression, based on an adapted Newton-Raphson method. Our proposal was driven by two goals: strong privacy protection and efficient computation. In below, we first provide a high-level overview of our framework; then we introduce the mathematical derivation underlying the method; later, we describe the detailed computations occurring at each stage of the framework and explain how data privacy is preserved thoroughly.

**Hybrid Architecture.** Our privacy-preserving method for performing $L_2$-regularized logistic regression features a hybrid architecture combining distributed (local) computing and centralized (secure) aggregation (Fig. 1). It is motivated by the observation that certain computations of model estimation could be decomposed per institution, resulting in institution-local computations and center-level aggregation. The careful partitioning and distributing of computations significantly accelerate the process compared with naïve centralized secure implementations of Newton-Raphson method, while still guaranteeing the same level of, if not stronger, privacy. Similar strategies of distributed computing have been explored in earlier works [6, 23] focused on other analytical tasks and prove successful in practice.

Without delving into technical details, we first introduce our framework as illustrated in Fig. 1. The framework (and the underlying iterative procedure) consists of two classes of computations: i) the *distributed phase* for computing institution-specific summary statistics locally at individual institutions, and ii) the *centralized phase* for securely aggregating and updating regression coefficient estimates. For each iteration, individual institutions independently compute their local summary statistics (i.e., denoted as *aggregates* in Fig. 1. These can be local gradient and Hessian matrix as introduced later) based on their own data, respectively. Such aggregates are then encrypted (via Shamir's secret-sharing [30] which will be explained later) and securely submitted to the Computation Centers (typically multiple independent Centers are designated to collectively hold the data for maximum security). The Computation Centers then collaborate to perform a series of secure data aggregation on the encrypted data, and perform the Newton-Raphson updating (Equation 2) to obtain a globally consistent $\beta$. In addition, model convergence checks will also be securely performed. The new $\beta$ (i.e., denoted as *adjustment* in Fig. 1) will then be redistributed to local instituions for the next iteration. The above process of distributed and centralized computing will proceed in iterations until model convergence criteria is satisfied.

**Newton-Raphson Method for $L_2$-regularized Logistic Regression.** Our framework (Fig. 1) leverages an adapted Newton-Raphson method for model estimation. Here we first demonstrate how the aforementioned Newton-Raphson method applies to $L_2$-regularized logistic regression. Then we identify the limitations of naïvely applying the method, which motivate us to derive a more efficient approach based on a hybrid architecture.

First, we reformulate the Newton-Raphson method (Equation 2) by defining a diagonal matrix $W$ as $w_{ii} = p_i(1 - p_i), \forall i = 1..N$, where $p_i$ corresponds to the probability estimate for the $i^{\text{th}}$ data record (i.e., a row) and $N$ denotes the total number of records. By expanding $\mathbf{H}(.)$ and $\mathbf{g}(.)$ for

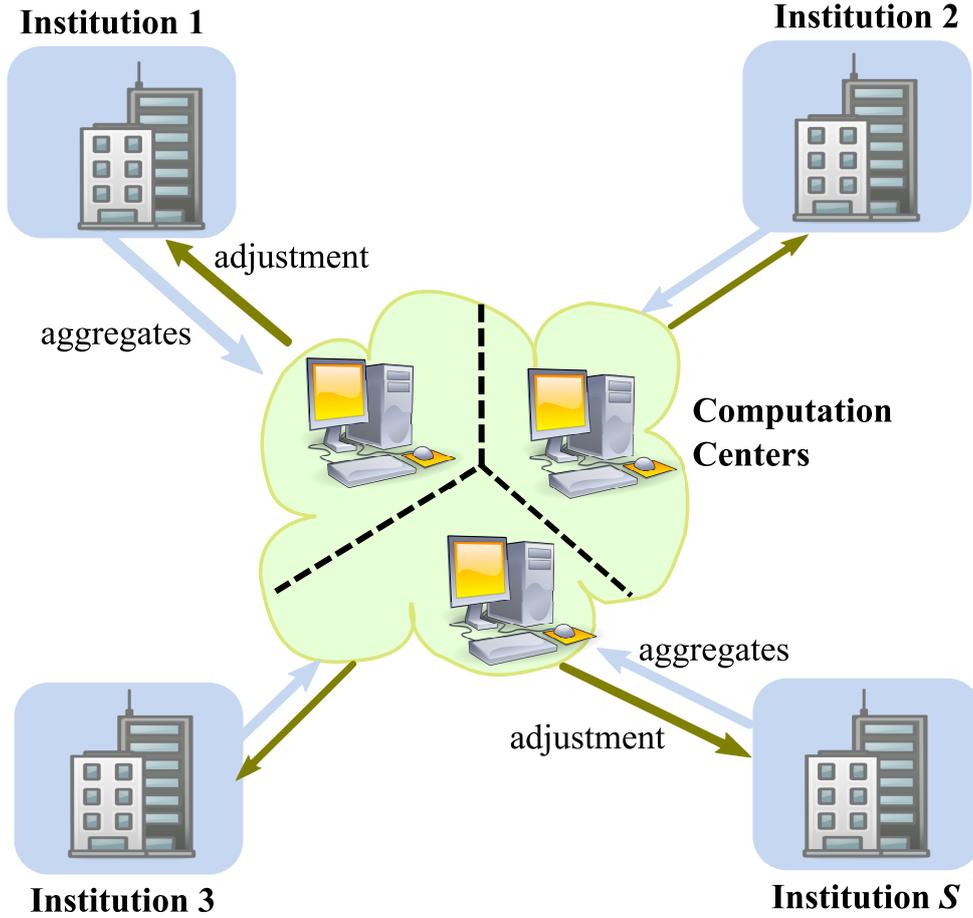

**Figure 1.** Overview of our secure framework for regularized logistic regression. Each institution (possessing private data) locally computes summary statistics from its own data, and submits encrypted aggregates following a strong Cryptographic scheme [30]. The Computation Centers securely aggregate the encryptions and conduct model estimation, from which the model adjustment feedback will be sent back as necessary. This iterative process continues until model convergence.

$L_2$-regularized logistic regression, Equation 2 becomes:

$$\boldsymbol{\beta}^{new} = \boldsymbol{\beta}^{old} + (\mathbf{XWX}^T + \lambda \mathbf{I})^{-1}(\sum_{i=1}^{N}(1-p_i)y_i\mathbf{x}_i - \lambda\boldsymbol{\beta}^{old}), \tag{3}$$

where $\mathbf{X}$ corresponds to the design matrix (i.e., covariates) of dimension $N \times d$, $\lambda$ is the regularization parameter for the $L_2$-norm (defined *a priori* or derived via cross-validation), and $\mathbf{I}$ denotes the identity matrix.

Traditionally, the aforementioned model estimation method (Equation 3) proceeds in a centralized fashion. This indicates that all individual-level raw data are pooled into one large (centralized) collection, on which the Hessian matrix and gradient are computed and the Newton-Raphson updating applied. Similar approach is also commonly pursued by the privacy-preserving data mining community (e.g., [37]).

We point out that such a centralized approach could suffer from severe computational inefficiency especially for large studies with privacy protection requirement. In particular, pooling raw data often results in datasets of large scale, on which secure computations can be prohibitively slow (if not infeasible) due to the complexity of supporting matrix operations in secure. Consequently, many alternative privacy-preserving proposals (e.g., [37]) do not seem practical especially for large studies. Such limitations have been illustrated in subsequent studies even on very simple analytical tasks [38].

**Distributed Model Estimation.** Observing the inefficiency of the centralized Newton-Raphson method, we intend to accelerate the process by carefully partitioning the computations to extract "safe" procedures that can be performed more efficiently without violating privacy. Such a solution leads to two anticipated benefits: First, the majority of computations could be supported without relying on expensive secure computation techniques; Second, careful partitioning of computations guarantees the same level of privacy as centralized secure alternatives. We point out it is increasingly the trend to leverage distributed computing for better computational efficiency in privacy-preserving frameworks [38]. The partitioning of Newton-Raphson method has proven successful on other simpler tasks [6] than ours.

To accelerate the Newton-Raphson method (Equation 3), we observe that the computations of $\mathbf{H}(.)$ and $\mathbf{g}(.)$ in Equation 2 can be decomposed, such that some sub-procedures can be performed locally at each institutions on their own respective data where privacy is not of concern. More formally, the per-institution decomposition of computations can be expressed as:

$$\mathbf{H}(\boldsymbol{\beta}) = -\sum_{i=1}^{N} w_{ii}(t)\mathbf{x}_i\mathbf{x}_i^T - \lambda\boldsymbol{\beta} = -\sum_{j=1}^{S} \underbrace{\overbrace{\sum_{i=1}^{N_j} w_{ii}(t)\mathbf{x}_i\mathbf{x}_i^T}^{\text{Per-institution } \mathbf{H}_j(\boldsymbol{\beta})} - \lambda\boldsymbol{\beta}}_{\text{All institutions}} \quad (4)$$

and

$$\mathbf{g}(\boldsymbol{\beta}) = \sum_{i=1}^{N}(1-p_i)y_i\mathbf{x}_i - \lambda\boldsymbol{\beta} = \sum_{j=1}^{S} \underbrace{\overbrace{\sum_{i=1}^{N_j}(1-p_i)y_i\mathbf{x}_i}^{\text{Per-institution } \mathbf{g}_j(\boldsymbol{\beta})} - \lambda\boldsymbol{\beta}}_{\text{All institutions}} \quad (5)$$

where $S$ denotes the total number of participating (distributed) institutions and $N_j$ denotes the total number of data records for Institution $j$ – it is easy to see that $N = \sum_{j=1}^{S} N_j$.

According to this decomposition, each institution can individually compute their local $\mathbf{H}_j(.)$ and $\mathbf{g}_j(.)$ on their respective data collections following their traditional practice. Later, the global Computation Centers only need to securely aggregate these (protected) intermediate results to derive the globally consistent $\mathbf{H}(.)$ and $\mathbf{g}(.)$, which would facilitate the Newton-Raphson algorithm.

In addition, the deviance test (for checking model convergence) [1] can also be decomposed similarly, since it depends on the log-likelihood which can be regarded as a series of sums.

$$Dev = -2\log L(\boldsymbol{\beta}) = -2 \underbrace{\sum_{j=1}^{S} \overbrace{\sum_{i=1}^{N_i} (y_i \log p_i + (1-y_i)\log(1-p_i))}^{\text{Per-institution } dev_j}}_{\text{All institutions}}, \qquad (6)$$

where $L(\boldsymbol{\beta})$ corresponds to the likelihood.

Based on the above intuition, we introduce a hybrid architecture for supporting $L_2$-regularized logistic regression. The framework features an iterative process composed of two types of computations: distributed (local) computation and centralized aggregation. In the following sections, we will describe these computations in greater detail.

**Distributed Computation.** The goal of the distributed computation phase is for local institutions to pre-compute their respective summary statistics. During this phase (Steps 3 through 8 in Algorithm 1), each participating institutions compute their local Hessian matrix $\mathbf{H_j}$ and gradient $\mathbf{g_j}$ (Equations 4 and 5) using their own data. Local deviance test $dev_j$ can also be computed similarly (Equation 6). Since each institution has complete ownership over their respective data and no data sharing is involved, such local computations naturally preserve privacy without requiring computationally-expensive Cryptographic protections.

Next, all intermediate summary statistics (e.g., $\mathbf{H_j}$, $\mathbf{g_j}$, $dev_j$) need to be synthesized and processed at the center level to obtain a globally fitting coefficient estimate. To prevent potential privacy inference attacks on aggregate information (partially summarized in [13, 25, 26]), we require each institution to obfuscate their local summaries prior to data submission (Step 7 in Algorithm 1) leveraging a strong protection mechanism known as Shamir's secret-sharing [30] (also introduced later). This mechanism ensures that all intermediate summary statistics (the "secrets") are split into multiple shares to be collectively held by many participants (e.g., one participant would possess only one piece of the secret). The actual content of the "secrets" can only be recovered if the majority of share-holding participants cooperate to decrypt. This way, even if there is collusion between a (minority) few of the secret-share holders, the system is still secure. For our use case, we designate independent Computation Centers to be share holders.

**Centralized Aggregation.** Once the distributed computation is completed, the subsequent phase of centralized computation should follow. As the first step, the Computation Centers will aggregate the respective (secret-share-protected) data submissions in a secure way. This process requires collaboration between the Centers who hold the "secrets". Once the globally adjusted $\mathbf{H}(.)$ and $\mathbf{g}(.)$ are derived, the Computation Centers will perform the Newton-Raphson updating on the $\boldsymbol{\beta}^{old}$ estimate and check for model convergence afterwards. If the model is still not converged, then the updated $\boldsymbol{\beta}^{new}$ estimate will be redistributed to local institutions to initiate the next iteration of running.

---
**Algorithm 1** Secure estimation of $L_2$-regularized logistic regression.
---
**Require:** Penalty parameter $\lambda$; Previous-iteration regression coefficients $\boldsymbol{\beta}^{old}$
**Ensure:** New regression coefficients $\boldsymbol{\beta}^{new}$
 1: **while** model not converged **do**
 2:    *[Local Institutions]*
 3:    **for** Institution j = 1 **to** S **do**
 4:       Compute Hessian matrix $\mathbf{H}_j$
 5:       Compute gradient $\mathbf{g}_j$
 6:       Compute deviance $dev_j$
 7:       Protect $\mathbf{H}_j$, $\mathbf{g}_j$, $dev_j$ via Shamir's secret-share
 8:    **end for**
 9:
10:    *[Computation Centers]*
11:    Securely aggregate: $\mathbf{H} = -\sum_{j=1}^{S} \mathbf{H}_j - \lambda \boldsymbol{\beta}^{old}$
12:    Securely aggregate: $\mathbf{g} = \sum_{j=1}^{S} \mathbf{g}_j - \lambda \boldsymbol{\beta}$
13:    Securely aggregate: $Dev = \sum_{j=1}^{S} dev_j$
14:
15:    Securely derive $\boldsymbol{\beta}^{new}$ via Newton-Raphson method
16:    Check for model convergence
17: **end while**
---

**Protecting Privacy**

The presented framework involves various types of data and computations, many of which are sensitive or quasi-sensitive. In this section, we analyze how privacy are preserved at each level.

**Privacy on Individual Data.** The hybrid architecture is designed in such a way that individual raw data are fully controlled by their owner institution only, and no individual-level data sharing is involved in any subsequent computations. As a result, individual-level privacy is maintained. We note that decoupling from raw individual data for privacy protection is a proven and widely pursued approach in methodological development in genetics and related fields, as evidenced by [6, 13].

**Privacy on Aggregate Data.** We observe that various inference attacks on privacy are only possible due to the disclosure of summary statistics. For instance, the genome-disease inference attack in [24] relies on certain genomic summaries of case/control groups; it has also been analyzed in [25–27] about the risks associated with disclosing summary statistics, such as covariance matrix, information matrix and score vector. Meanwhile, we note that aggregate data may also constitute confidential or proprietary information for some institutions and thus should be protected (a similar opinion was briefly communicated in [23]). This is not uncommon for joint studies in competitive scenarios, such as financial collaborations, healthcare quality comparisons, and association studies involving sensitive and rare diseases.

Specific to our task of regularized logistic regression (and logistic regression in general), the vulnerable summaries are the hessian and gradient, which collectively could result in inference attacks on private response variables and model recovery [13, 25, 26].

To prevent potential attacks or confidentiality breaches, our framework encrypts summary statistics from participating institutions (prior to data submission to Computation Centers) leveraging a strong Cryptographic mechanism known as Shamir's secret-sharing [30](to be introduced in the following section). The idea of protecting intermediate data have been explored before [23, 27, 38], however, mostly on simpler tasks (e.g., ridge linear regression, standard logistic regression, etc) than ours; in a more related work [23], summaries from distributed Newton method have been obfuscated, however, the protection is insufficient and easily vulnerable to collusion attacks as we will discuss later.

**Shamir's Secret-Sharing for Protecting Data.** In our protocol, we leverage Shamir's secret-sharing [30] to protect intermediate data (including summary statistics from institutions). The general idea underlying Shamir's secret sharing is that for a $t$-dimensional Cartesian plane, at least $t$ independent coordinate pairs are necessary to uniquely determine a polynomial curve. Formally, a $t$-out-of-$w$ secret-share scheme is defined as follows: we intend to protect a secret $m$ (e.g., certain institution-specific summary statistic in our case) such that the only way to successfully recover the secret is through cooperation of at least $t$ (i.e., the "threshold") share-holding participants (out of a total of $w$). To achieve the goal, we construct a random polynomial $q(x)$ of degree $(t-1)$ with the secret $m$ embedded (we point out that the calculations actually occur in a finite integer field. However, for presentation simplicity, we skip the technical details):

$$q(x) = m + \sum_{i=1}^{t-1} a_i x^i, \qquad (7)$$

where $m$ is the secret we want to protect, and $a_i$'s are randomly generated polynomial coefficients. Note that the polynomial itself will be kept secret.

In order to split and "share" the secret, we proceed to evaluate $q(x)$ and derive $t$ or more distinct values from the polynomial, yielding coordinate pairs $(1, q(1)), (2, q(2)), ..., (t, q(t)), ..., (w, q(w))$. Due to the inherent randomness in the specified polynomial, the coordinate pairs we obtain here are random and reveal nothing meaningful about the secret. These pairs, each of which constitutes a share of the secret, are then distributed to $t$ or more Computation Centers, respectively (i.e., each participant only receives one piece of the secret). Under this mechanism, we can claim that the secret is successfully protected, since no single Center or a limited few are capable of inferring anything about the polynomial or the embedded secret. When it is necessary to recover the original secret, $t$ or more share holders will collectively perform Lagrange polynomial interpolation [30] to uniquely determine the polynomial $q(x)$. The secret will naturally emerge by evaluating $q(0)$: $m = q(0)$. To facilitate complex data and computations in our framework, we have extended the scheme to support matrices and vectors.

**Privacy on Computations.** Since all data in our framework are in encrypted form, special care must be taken to support analytical procedures. Here we introduce several secure primitives for supporting necessary computations without decrypting intermediate data. We focus on secure addition and secure multiplication by a public value.

Secure addition is a fundamental building block for the central aggregation phase (Steps 11 through 13). Briefly, the primitive helps securely derive the sum $A + B$ without knowing the actual content of $A$ and $B$, since both of which are encrypted via Shamir's secret-sharing. As illustrated in Algorithm 2, the general idea of the secure addition primitive is to ask each share holders to locally aggregate original shares of the two secret addends in order to derive new shares, which will serve as the shares for their sum.

---
**Algorithm 2** Secure addition (aggregation).

---
**Require:** Secret-shared data $A$ and $B$ (among $w$ institutions)
**Ensure:** Sum $sum = A + B$ in secret-shared form (among $w$ institutions)
 1: **for** institution j := 1 **to** w **do**
 2:    *[At Institution j]*
 3:    Compute and store new share: $sum_j = A_j + B_j$
 4: **end for**

---

To show that the secure addition primitive is correct, we assume the (secret-sharing) polynomials to be $q_A(x), q_B(x)$, respectively, for the two secrets $A, B$. In other words: $A = q_A(0), B = q_B(0)$. Since both polynomials share the same covariates and degrees, we have: $q_A(0) + q_B(0) = (q_A + q_B)(0)$. This indicates that, the aggregated coordinate pairs satisfy the newly defined polynomial $(q_A + q_B)(.)$ and thus represent the new shares of the to-be-computed sum $A + B$.

Next, we show how secure multiplication-by-a-constant can be implemented, which is required by the Newton-Raphson method. In particular, we consider multiplying a secret value (in secret-shared form) by a known constant value. The primitive is surprisingly simple: share holders only need to locally multiply their shares (of the secret value) by the public constant to derive the new shares for the product of the two values. The proof for this method is straightforward, since multiplication by a constant can be reformulated as a series of secure additions by the secret value itself.

Note that in our current implementation, we take a pragmatic approach to security for better computational efficiency without degrading privacy. Specifically, the primary reason why protecting intermediate data is necessary in regularized logistic regression is due to privacy inference attacks [13, 25, 26]. Existing attacks rely on both hessian and gradient to be feasible. Our protection thus only needs to protect one of the summaries to prevent such attacks. This can lead to significant speedup as compared to an encrypting-all strategy and our privacy protection goal is still achieved. Extending our current implementation to a fully encrypted setting is also straightforward, as the additional secure primitives (e.g., secure matrix inversion) have already been demonstrated before [38].

Secure matrix inversion can be useful if we want to fully secure intermediate computations (e.g., inverting the Hessian matrix). Several existing secure solutions [38–40] serve as our reference that leverage methods such as LU-decomposition, Gaussian elimination, etc. Due to the focus of this work, we leave it as future extension.

Since none of our aforementioned primitives change the original Shamir's scheme, the information-theoretical security still holds in our protocol. Interested readers are kindly referred to relevant literature [41] for a detailed security proof.

**Generating synthetic data.** To allow for comprehensive evaluation on our framework, we also generate synthetic datasets (in addition to other real datasets as introduced later) according to Algorithm 3. We first generate coefficients and covariates at random (according to uniform and Gaussian distributions, respectively). Later, based on the calculated probabilities, we generate the response variables from the Bernoulli distribution. The resulting synthetic dataset is partitioned per institution.

## Results

We have implemented our privacy-preserving framework for $L_2$-regularized logistic regression. To validate our proposal, we performed extensive empirical evaluation on both synthetic and real-world studies. We report on the evaluations in terms of result accuracy, computational efficiency, as well as scalability to large studies.

**Algorithm 3** Generate synthetic data
---
**Require:** Covariate dimensionality $d$
**Ensure:** Covariates **X**, responses **y** (both partitioned per-institution)
 1: Generate coefficients $\boldsymbol{\beta} \in \mathbb{R}^d$ at random
 2: **for** institution j := 1 **to** S **do**
 3:     Generate covariates $\mathbf{cov}_j \in \mathbb{R}^{N_j \times (d-1)}$ from $\mathcal{N}(\mu, \sigma^2)$
 4:     Output concatenated covariates $\mathbf{X}_j = \begin{bmatrix} 1 & \mathbf{cov}_j \end{bmatrix} \in \mathbb{R}^{N_j \times d}$
 5:     Calculate probabilities $\mathbf{p}_j = 1/(1 + e^{-\boldsymbol{\beta}^T \mathbf{X}_j}) \in \mathbb{R}^{N_j}$
 6:     Generate and output response variables $\mathbf{y}_j \in \mathbb{R}^{N_j}$ from *Bernoulli*($\mathbf{p}_j$)
 7: **end for**
---

**Evaluation Datasets**

Included in our empirical evaluation are four studies, which represent a wide spectrum of applications from different domains and data scales. In specific,

- The **Synthetic dataset** is a large-scale dataset we generated at random according to Algorithm 3. It consists of 1 million records spanning 6 features from 6 institutions.

- The **Insurance dataset** [42] is a dataset from an insurance company with the goal of predicting users' insurance policy status based on socio-demographic features. It contains 9,822 records and 84 features, and we simulated 5 institutions by randomly partitioning the dataset horizontally.

- The **Parkinsons.Motor and Parkinsons.Total datasets** both relate to one dataset targeted for predicting parkinson's tele-monitoring quantities, with 5,875 samples spanning 20 features [43]. Since there are two distinct target predictions in the original dataset, we partition the dataset into two sub-studies which we denote as **Parkinsons.Motor** (for predicting motor UPDRS) and **Parkinsons.Total** (for total UPDRS). They share the same covariates but with different response variables. We randomly partitioned the records among 5 institutions.

**Regression Result Accuracy**

The first question we consider in validating our framework is whether the regression result is accurate and reliable. To answer this question, we compare our estimated regression coefficients with that obtained from standard software packages. As illustrated in Fig. 2, our framework yields identical results to the expected ground truth across all evaluations (with correlation $R^2 = 1.00$). The result accuracy is also evidenced by the mathematical proof explained earlier, where we have shown that our distributed model estimation method follows an exact derivation and no approximation is involved in the secure computation procedures.

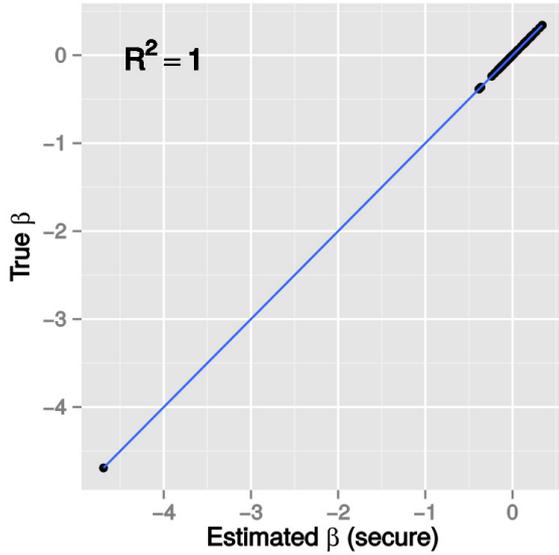

(a) Insurance dataset.

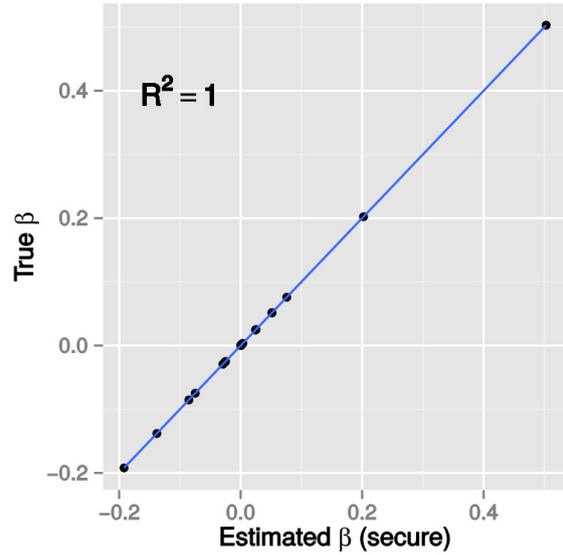

(b) Parkinsons.Motor dataset.

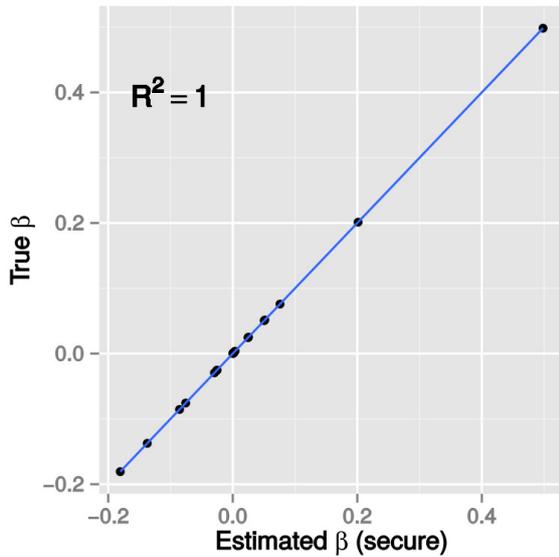

(c) Parkinsons.Total dataset.

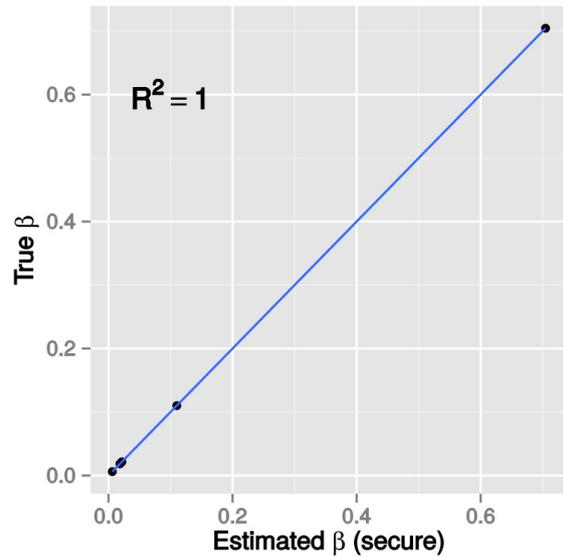

(d) Synthetic dataset.

**Figure 2.** Model accuracy of our securely estimated $\beta$ against the gold standard for four evaluation datasets. As illustrated, the regression coefficients estimated via our secure framework are identical to the gold standards, with correlation $R^2 = 1.00$.

**Running Time**

We implemented the prototype in R and Scala, a Java Virtual Machine-based programming language. Experiments were performed on a quad-core computer with 2.4GHz CPU and 8GB memory, running Ubuntu 13.04. To eliminate network latency effects, we simulated distributed computing nodes on a single computer and report the network data exchanged. We performed each experiment several times and reported the mean of the running time.

Empirical evaluation indicates that our protocol is highly efficient, as demonstrated in Table 1. For datasets with as many as 1 million records, our protocol completed in less than 12 seconds. For datasets of more modest sizes as typically found in everyday applications, our protocol took only around 2 ∼ 4 seconds or less.

Since our framework is focused on a novel analytical application that is not addressed in the privacy/security domain, technically we do not have any alternatives to compare against. We do however, try to provide brief comparisons against similar secure approaches in related problems – mostly from linear (ridge) regression which also considered regularization. Our evaluation indicate that our protocol is more efficient than other related secure proposals (even though they focused on much simpler regression models). For instance, as a rough comparison, secure linear regression in [39] on 51,016 samples with 22 covariates took two days. Our framework is also competitive compared with the state-of-the-art secure solution for ridge (linear) regression [38] (a much simpler model), which took 55 seconds on a smaller-scale Insurance dataset (with only 14 features). We do acknowledgment that such comparisons are not very fair, as our proposal solves a different and more complicated regression model; also some alternatives implemented additional features. Nevertheless, the results demonstrate that our secure framework for regularized logistic regression is efficient and competitive.

**Table 1. Computational efficiency on evaluation datasets.**

| Dataset | Insurance | Parkinsons.Motor | Parkinsons.Total | Synthetic |
|---|---|---|---|---|
| # samples | 9,822 | 5,875 | 5,875 | 1,000,000 |
| # features | 84 | 20 | 20 | 6 |
| # iterations | 8 | 6 | 6 | 6 |
| Central runtime (S) | 0.42 | 0.264 | 0.236 | 0.076 |
| Total runtime (S) | 3.77 | 2.017 | 2.352 | 12.76 |
| Data transmitted (MB) | 80 | 492 | 492 | 612 |

Overall, the repeated Newton-Raphson process converged within a limited number of iterations, as evidenced by Fig. 3. Across all evaluation datasets, the models converged within 6 ∼ 8 iterations. We set convergence criteria to be $10^{-10}$. Also, the amount of data to be exchanged during computation is also modest. As an example, for the Synthetic dataset with 1 million records, only around 612 megabytes of data are transmitted over the network.

To further demonstrate the efficiency of our method, we report time efficiency of its major procedures (i.e., the central phase and the total runtime) in Table 1. We emphasize that the vast majority of runtime is spent at individual local institutions (on ordinary computations), and secure computation at the Computation Centers only consumes around 11.14%, 13.09%, 10.03%, and 0.60% of the total time for the datasets evaluated, respectively.

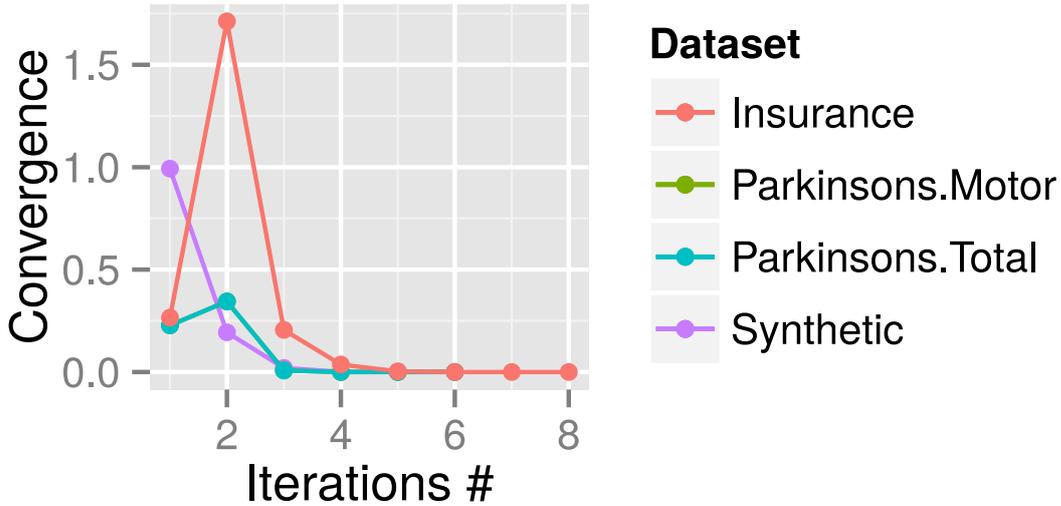

**Figure 3.** Model convergence (i.e., deviance) for all datasets (deviance smaller than the threshold indicates convergence). All models converged within 6 ∼ 8 iterations. Note that the convergence scores for the Parkinsons.Motor and Parkinsons.Total studies are so similar that they overlap on the plot.

**Scalability to Large Studies**

With the advent of the big data era, large-scale collaborative studies are becoming ubiquitous in many domains. A few notable examples include the International Cancer Genome Consortium [14], the Patient-Centered Outcomes Research Institute (PCORI) [44], and financial systematic risk protection [45].

To meet the demand of large-scale cross-institution studies, we also demonstrate the scalability of our framework. Since regression accuracy is not affected by the increase of participating institutions, we mainly focus on evaluating the running time. To do so, we simulated studies with up to 100 institutions, and reported the results in Figure 4 (we simplified the scenario by assuming that each institution contributes 10000 records. So in fact, our evaluation reflects the running time affected by the increase of both the number of institutions and the total number of data records).

It can be seen that the total time is always between 3.0 ∼ 3.3 seconds, exhibiting minimal fluctuation as the number of participating institutions increases. This is especially the case for the secure-computation-based centralized phase, which consistently takes only around 0.088 seconds.

Such a trend is well explained from a theoretical perspective (as evident in the computation details in Algorithm 1, as individual institutions perform their local (distributed) computations simultaneously without interacting with (or waiting for) other participants. As a result, local computations are relatively stable from the change. The increase of the number of institutions does slightly influence the centralized aggregation of institution-level summary statistics, as more summaries need to be transmitted and aggregated. But the effect is minimal, since the summary data size is relatively small and the majority of computations for aggregating secret shares occur locally at each Computation Center (as explained earlier regarding secure addition and multiplication).

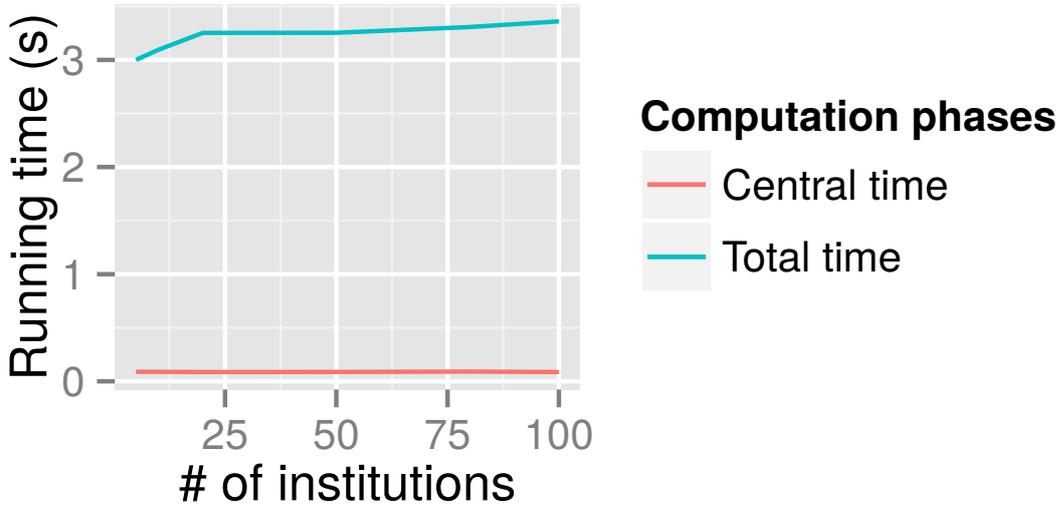

**Figure 4.** Running time (in seconds) for the central phase and total computation respectively, as the number of participating institutions increases. Negligible time fluctuation is present, especially for the central (secure) computation.

Overall, the evaluation has demonstrated that our secure framework could support large-scale studies with hundreds of institutions and millions of data records.

## Discussion

While the prototype implementation has already demonstrated impressive efficiency, we point out that further speed-ups can be obtained for production systems. For instance, local data can be cached in computer memory to greatly streamline and accelerate subsequent iterations of computations; further acceleration can be gained locally by adopting high-performance programming languages (e.g., C/C++) and libraries (e.g., BLAS/LAPACK [46]); as for the central computation, it can also greatly benefit from multi-core parallelism, since many secure operations can be parallelized naturally. In addition, the cryptography community continues to improve efficiency of secure primitives which could be useful to us in future. In addition to Shamir's secret-sharing we used here [30], there are also several alternative schemes that prove to be useful on many tasks, such as Paillier encryption and Yao's garbled circuit (as used by [13, 27, 38]). Due to space constraint, we intend to explore other potential schemes for related tasks in future.

There have been various alternative proposals for protecting privacy while supporting regression analysis. Most of them only focused on much simpler regression models, such as linear (ridge) regression, or standard logistic regression. And typically there is no or only weak protection over summary statistics. Here we primarily review cryptography-related approaches, which are directly relevant to our proposal. For instance, a privacy-preserving method was proposed for (linear) ridge regression [38], which directly solves the linear system in secure centrally. Other secure solutions [37, 39] for linear or logistic regression relied on some expensive secure primitives and approximations, which add significant computational overhead and do not seem scalable to large sample sizes. Increasingly, distributed-architecture-based solutions [6, 13, 23, 27] emerged as very promising solutions for linear/logistic regression and related analytics. However, none of them focused on regularized regression which is a more widely used model in practice. Many related proposals [6] directly expose summary data from model fitting, leading to serious privacy concerns over inference attacks on intermediate data [13, 25–27]. While others started to think about protecting institutional summary information, the protections seem rather weak. For instance, the obfuscation in [23] is vulnerable to collusion attacks by the center (who generates the randomization noise) and any of the institutions, causing single points of failure/breach.

Our framework demonstrated here for regularized logistic regression differentiates in several aspects. Firstly, we focus on an important and (more) widely-used statistical model that has not been addressed by the security/privacy community. While there is recent privacy-preserving work [38] specifically targeted for ridge (linear) regression (i.e., $L_2$-regularization), it focused on a much simpler regression model (i.e., linear regression) and the model estimation process is completely different from regularized logistic regression. None of the other related works considered regularization, despite its wide adoption and popularity in various application domains as well as methodological development in statistics and machine learning. Secondly, for efficient model estimation on regularized logistic regression, we adapted a distributed Newton method that previously has only been validated on simpler analytical models [6]. The distributed process makes our secure protocol for regularized logistic regression highly efficient compared to a straightforward centralized implementation. Thirdly, we protect intermediate data and computations with stronger cryptographic schemes [30], providing strong security guarantees thanks to decentralization of trust while still allowing for efficient and flexible computation. Ours is the first to safeguard intermediate data from regularized logistic regression. Among the two closely related works, [6] failed to provide any protection over summaries; And [23] had very weak protection as discussed earlier. Lastly, our model does not involve approximation or artificial perturbation (contrary to solutions based on classical *k*-anonymity [20] or differential privacy [21]) on the data or computations, thus maintaining accuracy of the predictive model.

**Application Scenarios**

We believe the proposed privacy-preserving framework is applicable to a wide range of domains where the privacy/confidentiality of study participants and/or institutions is of concern. Here we briefly describe a few representative application scenarios.

**Genetic and Biomedical Studies.** Genetic studies have enjoyed continued investigation efforts with the ultimate goal of uncovering connections between genes and human traits (e.g., diseases). Regularized logistic regression is an increasingly important tool for related applications, including for genomic selection [28, 47], gene-gene interactions [4], GWAS [34], etc. Other biomedical studies such as prediction of adverse drug reactions [48] are also potential application domains.

Many such studies rely on large-scale data sharing across institutions, while at the same time, many such data involve sensitive data such as genome information, or participant phenotypes [13]. We envision that our framework can provide an automated and privacy-preserving solution for supporting such collaborative investigations.

**Analytics for Smart Grid.** Smart electrical grid is a transformative technology that provides detailed data pertaining to the monitoring and management of energy consumption of individual households. Data sharing and analytics on such data have raised serious privacy concerns from both everyday consumers and governmental regulators [49] due to various privacy inference attacks on energy monitoring data. We believe that our distributed-computing-based technology can support some useful analytics on smart grid data, such that household privacy could be maintained.

**Large-scale Network Analysis.** Many important innovations involve analysis of social network data, such as [8, 50, 51]. These include anomaly detection, novel discoveries in online social networks (such as personalization and link prediction), etc. Social networks data often involve person-level private information, making them inappropriate to share across institutions in large collaborative studies. Our framework could serve the purpose by allowing for joint network analysis without disclosing private information.

# Conclusion

In this work, we focused on privacy protection for regularized logistic regression, a widely-used statistical model in various domains. To make the model efficient in a secure setting, we adapted a distributed method for model estimation. To further enhance privacy and prevent inference attacks over intermediate data during model estimation, we introduced strong cryptographic protections. These lead to an efficient framework for supporting regularized logistic regression across institutions while guaranteeing strong privacy both for individual study participants and institutions. Extensive empirical evaluations have demonstrated the efficacy of the framework in guaranteeing privacy with modest computational overhead. We hope that careful implementation of our framework could enable a wider range of cross-institution joint analytics, which would otherwise be impossible due to privacy or confidentiality concerns.

## Acknowledgments

We thank You Chen and Bradley A. Malin for helpful discussions.